\documentclass{llncs}
\usepackage{amsfonts}
\usepackage{amsmath,amssymb,amscd,epsfig,amsfonts,rotating}
\usepackage{graphicx}
\usepackage{epsfig,epstopdf}
\usepackage{multirow}
\usepackage{booktabs}
\usepackage{url}
\usepackage{xcolor, colortbl}
\usepackage{tikz}
\usepackage{makeidx}  
\usetikzlibrary{arrows}
\usetikzlibrary{fit}
\usepackage{algorithm}
\usepackage{algpseudocode}

\newcommand{\f}{\scriptsize}

\newcommand{\bs}{\boldsymbol}
\newcommand{\ft}{\texttt}
\newcommand{\ttt}{\texttt}
\newcommand{\tsf}{\textsf}
\newcommand{\tweb}{\text{c}} 
\newcommand{\tads}{\text{r}} 
\newcommand{\td}{\text{d}}

\begin{document}
\title{Implicit Look-alike Modelling in Display Ads}
\subtitle{-- Transfer Collaborative Filtering to CTR Estimation}

\author{Weinan Zhang\inst{1,2} \and Lingxi Chen\inst{1} \and Jun Wang\inst{1,2}}

\authorrunning{Weinan Zhang et al.} 

\institute{$^1$University College London, $^2$MediaGamma Limited\\
London, United Kingdom\\
\email{\{w.zhang,lingxi.chen,j.wang\}@cs.ucl.ac.uk}}

\maketitle

\begin{abstract}
User behaviour targeting is essential in online advertising. Compared with sponsored search keyword targeting and contextual advertising page content targeting, user behaviour targeting builds users' interest profiles via tracking their online behaviour and then delivers the relevant ads according to each user's interest, which leads to higher targeting accuracy and thus more improved advertising performance. The current user profiling methods include building keywords and topic tags or mapping users onto a hierarchical taxonomy. However, to our knowledge, there is no previous work that explicitly investigates the user online visits similarity and incorporates such similarity into their ad response prediction. In this work, we propose a general framework which learns the user profiles based on their online browsing behaviour, and transfers the learned knowledge onto prediction of their ad response. Technically, we propose a transfer learning model based on the probabilistic latent factor graphic models, where the users' ad response profiles are generated from their online browsing profiles. The large-scale experiments based on real-world data demonstrate significant improvement of our solution over some strong baselines.
\end{abstract}

\section{Introduction}
Targeting technologies have been widely adopted in various online advertising paradigms during the recent decade. According to the Internet advertising revenue report from IAB in 2014 \cite{iab2014report}, 51\% online advertising budget is spent on sponsored search (search keywords targeting) and contextual advertising (page content targeting), while 39\% is spent on display advertising (user demographics and behaviour targeting), and the left 10\% is spent on other ad formats like classifieds. With the rise of ad exchanges \cite{muthukrishnan2009ad} and mobile advertising, user behaviour targeting has now become essential in online advertising.

Compared with sponsored search or contextual advertising, user behaviour targeting \emph{explicitly} builds the user profiles and detects their interest segments via tracking their online behaviour, such as browsing history, search keywords and ad clicks etc. Based on user profiles, the advertisers can detect the users with similar interests to the known customers and then deliver the relevant ads to them. Such technology is referred as \emph{look-alike modelling} \cite{mangalampalli2011feature}, which efficiently provides higher targeting accuracy and thus brings more customers to the advertisers \cite{yan2009much}. The current user profiling methods include building keyword and topic distributions \cite{ahmed2011scalable} or clustering users onto a (hierarchical) taxonomy \cite{yan2009much}. Normally, these inferred user interest segments are then used as target restriction rules or as features leveraged in predicting users' ad response \cite{zhang2014real}.

However, the two-stage profiling-and-targeting mechanism is not optimal (despite its advantages of explainability). First, there is no flexible relationship between the inferred tags or categories. Two potentially correlated interest segments are regarded as separated and independent ones. For example, the users who like cars tend to love sports as well, but these two segments are totally separated in the user targeting system. Second, the first stage, i.e., the user interest segments building, is performed independently and with little attention of its latter use of ad response prediction \cite{yan2009much,dalessandro2014scalable}, which is suboptimal. Third, the effective tag system or taxonomy structure could evolve over time, which makes it much difficult to update them.

In this paper, we propose a novel framework to \emph{implicitly} and \emph{jointly} learn the users' profiles on both the general web browsing behaviours and the ad response behaviours. Specifically,  (i) Instead of building explicit and fixed tag system or taxonomy, we propose to directly map each user, webpage and ad into a latent space where the shape of the mapping is automatically learned. (ii) The users' profiles on general browsing and ad response behaviour are jointly learned based on the heterogeneous data from these two scenarios (or tasks). (iii) With a maximum a posteriori framework, the knowledge from the user browsing behaviour similarity can be naturally transferred to their ad response behaviour modelling, which in turn makes an improvement over the prediction of the users' ad response. For instance, our model could automatically discover that the users with the common behaviour on \ttt{www.bbc.co.uk/sport} will tend to click automobile ads. Due to its implicit nature, we call the proposed model \emph{implicit look-alike modelling}.

Comprehensive experiments on a real-world large-scale dataset from a commercial display ad platform demonstrate the effectiveness of our proposed model and its superiority over other strong baselines. Additionally, with our model, it is straightforward to analyse the relationship between different features and which features are critical and cost-effective when performing transfer learning.


%

\section{Related Work}
\textbf{Ad Response Prediction} aims at predicting the probability that a specific user will respond (e.g., click) to an ad in a given context \cite{chapelle2014modeling,mcafee2011design}. Such context can be either a search keyword \cite{graepel2010web}, webpage content \cite{broder2007semantic}, or other kinds of real-time information related to the underlying user \cite{yuan2013real}. From the modelling perspective, many user response prediction solutions are based on linear models, such as logistic regression \cite{Richardson2007a,lee2012estimating} and Bayesian probit regression \cite{graepel2010web}. Despite the advantage of high learning efficiency, these linear models suffer from the lack of feature interaction and combination \cite{he2014practical}. Thus non-linear models such as tree models \cite{he2014practical} and latent vector models \cite{yan2014coupled,oentaryo2014predicting} are proposed to catch the data non-linearity and interactions between features. Recently the authors in \cite{juan2014idiot} proposed to first learn combination features from gradient boosting decision trees (GBDT) and, based on the tree leaves as features, learn a factorisation machine (FM) \cite{rendle2010factorization} to build feature interactions to improve ad click prediction performance.


\textbf{Collaborative Filtering (CF)} on the other hand is a technique for personalised recommendation \cite{schafer2007collaborative}. Instead of exploring content features, it learns the user or/and item similarity based on their interactions. Besides the user(item)-based approaches \cite{sarwar2001item,wang2006unifying}, latent factor models, such as probabilistic latent semantic analysis \cite{hofmann2003collaborative}, matrix factorisation \cite{koren2009matrix} and factorisation machines \cite{rendle2010factorization}, are widely used model-based approaches. The key idea of the latent factor models is to learn a low-dimensional vector representation of each user and item to catch the observed user-item interaction patterns. Such latent factors have good generalisation and can be leveraged to predict the users' preference on unobserved items \cite{koren2009matrix}. In this paper, we explore latent models of collaborative filtering to model user browsing patterns and use them to infer users' ad click behaviour.

\textbf{Transfer Learning} deals with the learning problem where the learning data of the target task is expensive to get, or easily outdated, via transferring the knowledge learned from other tasks \cite{pan2010survey}. It has been proven to work on a variety of problems such as classification \cite{dai2007transferring}, regression \cite{liao2005logistic} and collaborative filtering \cite{li2009transfer}. Different from multi-task learning, where the data from different tasks are assumed to drawn from the same distribution \cite{taylor2009transfer}, transfer learning methods may allow for arbitrary source and target tasks. In online advertising field, the authors in a recent work \cite{dalessandro2014scalable} proposed a transfer learning scheme based on logistic regression prediction models, where the parameters of ad click prediction model were restricted with a regularisation term from the ones of user web browsing prediction model. In this paper, we consider it as one of the baselines.


\section{Implicit Look-alike Modelling}
In performance-driven online advertising, we commonly have two types of observations about underlying user behaviours: one from their browsing behaviours (the interaction with webpages) and one from their ad responses, e.g., conversions or clicks, towards display ads (the interactions with the ads) \cite{dalessandro2014scalable}. There are two predictions tasks for understanding the users:

\begin{itemize}
\item \textbf{Web Browsing Prediction (CF Task).} Each user's online browsing behaviour is logged as a list containing previously visited publishers (domains or URLs). A common task of using the data is to leverage collaborative filtering (CF) \cite{wang2006unifying,rendle2010factorization} to infer the user's profile, which is then used to predict whether the user is interested in visiting any given new publisher. Formally, we denote the dataset for CF as $D^{\tweb}$ and an observation is denoted as $(\bs{x}^{\tweb}, y^{\tweb}) \in D^{\tweb}$, where $\bs{x}^{\tweb}$ is a feature vector containing the attributes from the user and the publisher and $y^{\tweb}$ is the binary label indicating whether the user visits the publisher or not.

\item \textbf{Ad Response Prediction (CTR Task).} Each user's online ad feedback behaviour is logged as a list of pairs of ad impression events and their corresponding feedbacks (e.g., click or not). The task is to build a click-through rate (CTR) prediction model \cite{chapelle2013simple} to estimate how likely it is that the user will click a specific ad impression in the future. Each ad impression event consists of various information, such as user data (cookie ID, location, time, device, browser, OS etc.), publisher data (domain, URL, ad slot position etc.), and advertiser data (ad creative, creative size, campaign etc.). Mathematically, we denote the ad CTR dataset as $D^{\tads}$ and its data instance as $(\bs{x}^{\tads}, y^{\tads})$, where $\bs{x}^{\tads}$ is a feature vector and $y^{\tads}$ is the binary label indicating whether the user clicks a given ad or not.

\end{itemize}

This paper focuses on the latter task: ad response prediction. We, however, observe that although they are different prediction tasks, the two tasks share a large proportion of users, publishers and their features. We can thus build a user-publisher interest model jointly from the two tasks. Typically we have a large number of observations about user browsing behaviours and we can use the knowledge learned from publisher CF recommendation to help infer display advertising CTR estimation.

\subsection{The Joint Conditional Likelihood}
In our solution, the prediction models on CF task and CTR task are learned jointly. Specifically, we build a joint data discrimination framework. We denote $\Theta$ as the parameter set of the joint model with prior $P(\Theta)$, and the \emph{conditional} likelihood of an observed data instance is the probability of predicting the correct binary label given the features $P(y|\bs{x}; \Theta)$. As such, the conditional likelihood of the two datasets are $\prod_{(\bs{x}^{\tweb}, y^{\tweb})\in D^{\tweb}} P(y^{\tweb} | \bs{x}^{\tweb}; \Theta)$ and $\prod_{(\bs{x}^{\tads}, y^{\tads})\in D^{\tads}} P(y^{\tads} | \bs{x}^{\tads}; \Theta)$. Maximising a posteriori (MAP) estimation gives
\begin{align}
\hat \Theta = \max_{\Theta} P(\Theta) \prod_{(\bs{x}^{\tweb}, y^{\tweb})\in D^{\tweb}} P(y^{\tweb} | \bs{x}^{\tweb}; \Theta) \prod_{(\bs{x}^{\tads}, y^{\tads})\in D^{\tads}} P(y^{\tads} | \bs{x}^{\tads}; \Theta) \label{eq:map}.
\end{align}
Just like most solutions on CF recommendation \cite{koren2009matrix,hofmann2003collaborative} and CTR estimation \cite{Richardson2007a,lee2012estimating}, in this discriminative framework, $\Theta$ is only concerned with the mapping from the features to the labels (the conditional probabilities) rather than modelling the prior distribution of features \cite{jebara2012machine}.

The details of the conditional likelihood $P(y^{\tweb} | \bs{x}^{\tweb}; \Theta)$, $P(y^{\tads} | \bs{x}^{\tads}; \Theta)$ and the parameter prior $P(\Theta)$ will be discussed in the latter subsections.


\subsection{CF Prediction}
For the CF task, we use a factorisation machine \cite{rendle2010factorization} as our prediction model. We further define the features $\bs{x}^{\tweb} \equiv (\bs{x}^u, \bs{x}^p)$, where $\bs{x}^u \equiv \{x_i^u\}$ is the set of features for a user and $\bs{x}^p \equiv \{x_j^p\}$ is the set of features for a publisher\footnote{All the features studied in our work are one-hot encoded binary features.}.
The parameter $\Theta \equiv (w_0^{\tweb},\bs{w}^{\tweb},\bs{V}^{\tweb})$, where $w_0^{\tweb}\in \mathbb{R}$ is the global bias term and $\bs{w}^{\tweb}\in \mathbb{R}^{I^{\tweb}+J^{\tweb}}$ is the weight vector of the $I^{\tweb}$-dimensional user features and $J^{\tweb}$-dimensional publisher features. Each user feature $x_i^u$ or publisher feature $x_j^p$ is associated with a $K$-dimensional latent vector $\bs{v}_i^{\tweb}$ or $\bs{v}_j^{\tweb}$. Thus $\bs{V}^{\tweb}\in \mathbb{R}^{(I^{\tweb}+J^{\tweb}) \times K}$.

With such setting, the conditional probability for CF in Eq. (\ref{eq:map}) can be reformulated as:
\begin{align}
\prod_{(\bs{x}^{\tweb},y^{\tweb})\in D^{\tweb}} P(y^{\tweb}|\bs{x}^{\tweb};\Theta)
=\prod_{(\bs{x}^u,\bs{x}^p,y^{\tweb})\in D^{\tweb}} P(y^{\tweb}|\bs{x}^u,\bs{x}^p;w_0^{\tweb},\bs{w}^{\tweb},\bs{V}^{\tweb}).\label{eq:web-likelihood}
\end{align}
Let $\hat{y}_{u,p}^{\tweb}$ be the predicted probability of whether user $u$ will be interested in visiting publisher $p$. With the FM model, the likelihood of observing the label $y^{\tweb}$ given the features $(\bs{x}^u, \bs{x}^p)$ and parameters is
\begin{align}
P(y^{\tweb}|\bs{x}^u,\bs{x}^p;w_0^{\tweb},\bs{w}^{\tweb},\bs{V}^{\tweb}) = (\hat{y}_{u,p}^{\tweb})^{y^{\tweb}} \cdot (1 - \hat{y}_{u,p}^{\tweb})^{(1 - y^{\tweb})}, \label{eq:web-pred}
\end{align}
where the prediction $\hat{y}_{u,p}^{\tweb}$ is given by an FM with a logistic function:
\begin{align}
\hat{y}_{u,p}^{\tweb} &= \sigma \Big( w_0^{\tweb} + \sum_{i} w_i^{\tweb} x_i^u + \sum_{j} w_j^{\tweb} x_j^p +\sum_i\sum_j \langle \bs{v}_i^{\tweb}, \bs{v}_j^{\tweb}\rangle x_i^u x_j^p \Big),\label{eq:fm-cf}
\end{align}
where $\sigma(x) = 1 / (1 + e^{-x})$ is the sigmoid function and $\langle\cdot,\cdot\rangle$ is the inner product of two vectors: $\langle \bs{v}_i,\bs{v}_j \rangle \equiv \sum_{f=1}^K v_{i,f} \cdot v_{j,f}$,
which models the interaction between a user feature $i$ and a publisher feature $j$.


\subsection{CTR Task Prediction Model}
For a data instance $(\bs{x}^{\tads}, y^{\tads})$ in ad CTR task dataset $D^{\tads}$, its features $\bs{x}^{\tads} \equiv (\bs{x}^u, \bs{x}^p, \bs{x}^a)$ can be divided into three categories: the user features $\bs{x}^u$ (cookie, location, time, device, browser, OS, etc.), the publisher features $\bs{x}^p$ (domain, URL etc.), and the ad features $\bs{x}^a$ (ad slot position, ad creative, creative size, campaign, etc.). Each feature has potential influence to another one in a different category. For example, a mobile phone user might prefer square-sized ads instead of banner ads; users would like to click the ad on the sport websites during the afternoon etc.

By the same token as CF prediction, we leverage factorisation machine and the model parameter thus is $\Theta \equiv (w_0^{\tads},\bs{w}^{\tads},\bs{V}^{\tads})$. Specifically, $x_l^a$ is one of the $L^{\tads}$-dimensional ad features $\bs{x}^a$, $w_l^{\tads}$ is the corresponding bias weight for the feature, and the feature is also associated with a $K$-dimensional latent vector $\bs{v}_l^{\tads}$. Thus $\bs{V}^{\tads}\in \mathbb{R}^{(I^{\tads} + J^{\tads} + L^{\tads}) \times K}$. Similar to CF task, the CTR data likelihood is:
\begin{align}
\prod_{(\bs{x}^{\tads},y^{\tads})\in D^{\tads}} P(y^{\tads}|\bs{x}^{\tads};\Theta)
= \prod_{(\bs{x}^u,\bs{x}^p,\bs{x}^a,y^{\tads})\in D^{\tads}} P(y^{\tads}|\bs{x}^u,\bs{x}^p,\bs{x}^a;w_0^{\tads},\bs{w}^{\tads},\bs{V}^{\tads}).
\end{align}

Then the factorisation machine with logistic activation function $\sigma(\cdot)$ is adopted to model the click probability over a specific ad impression:
\begin{align}
P(y^{\tads}|\bs{x}^u,\bs{x}^p,\bs{x}^a;w_0^{\tads},\bs{w}^{\tads},\bs{V}^{\tads}) = (\hat{y}_{u,p,a}^{\tads})^{y^{\tads}} + (1 - \hat{y}_{u,p,a}^{\tads})^{(1 - y^{\tads})}, \label{eq:ads-pred}
\end{align}
where $\hat{y}_{u,p,a}^{\tads}$ is modelled by interactions among 3-side features
\begin{align}
\hat{y}_{u,p,a}^{\tads} = \sigma \Big( &w_0^{\tads} +\sum_{i} w_i^{\tads} x_i^u + \sum_{j} w_j^{\tads} x_j^p + \sum_{l} w_l^{\tads} x_l^a + \label{eq:fm-ctr}\\
& \sum_i\sum_j \langle \bs{v}_i^{\tads}, \bs{v}_j^{\tads}\rangle x_i^u x_j^p + \sum_i\sum_l \langle \bs{v}_i^{\tads}, \bs{v}_l^{\tads}\rangle x_i^u x_l^a + \sum_j\sum_l \langle \bs{v}_j^{\tads}, \bs{v}_l^{\tads}\rangle x_j^p x_l^a \Big). \nonumber
\end{align}

\subsection{Dual-Task Bridge}
To model the dependency between the two tasks, the weights of the user features and publisher features in CTR task are assumed to be generated from the counterparts in CF task (as a prior):
\begin{align}
\bs{w}^{\tads} \sim \mathcal{N}(\bs{w}^{\tweb}, \sigma_{\bs{w}^\td}^2 \bs{I}),
\end{align}
where $\sigma_{\bs{w}^\td}^2$ is the assumed variance of the Gaussian generation process between each pair of feature weights of CF and CTR tasks and the weight generation is assumed to be independent across features.
Similarly, the latent vectors of CTR task are assumed to be generated from the counterparts of CF task:
\begin{align}
\bs{v}^{\tads}_i \sim \mathcal{N}(\bs{v}^{\tweb}_i, \sigma_{\bs{V}^\td}^2 \bs{I})
\end{align}
where $i$ is the index of a user or publisher feature; $\sigma_{\bs{V}^\td}^2$ is defined similarly.


The rational behind the above bridging model is that the users' interest towards webpage content is relatively general and the displayed ad can be regarded as a special kind of webpage content. One can infer user interests from their browsing behaviours, while their interests on commercial ads can be regarded as a modification or derivative from the learned general interests.

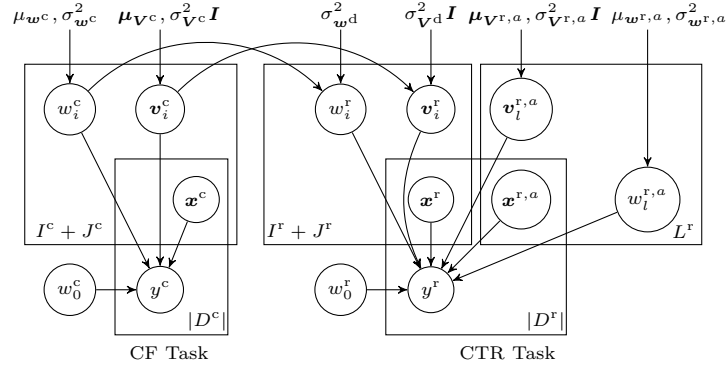
\begin{figure}[t]\vspace{-30pt}
\f
\begin{center}
\begin{tikzpicture}[->,>=stealth',scale=0.6]
  \node at (-1.8,-1.9) [circle] {$\bs{\mu}_{\bs{V}^\tweb},\sigma_{\bs{V}^\tweb}^2 \bs{I}$};
  \node at (-4.3,-1.9) [circle] {$\mu_{\bs{w}^\tweb},\sigma_{\bs{w}^\tweb}^2$};
  \node at (-2,-2) [circle] (uvl) {};
  \node at (-4,-2) [circle] (uwl) {};
  \node at (-2,-4) [circle,draw] (Vl) {$\bs{v}^{\tweb}_i$};
  \node at (-4,-4) [circle,draw] (wl) {$w^{\tweb}_i$};
  \node at (-1.2,-6) [circle,draw] (xl) {$\bs{x}^{\tweb}$};
  \node at (-4,-8) [circle,draw] (w0l) {$w_0^{\tweb}$};
  \node at (-2,-8) [circle,draw] (yl) {$y^{\tweb}$};
  \node at (4,-1.9) [circle] {$\sigma_{\bs{V}^{\td}}^2 \bs{I}$};
  \node at (2,-1.9) [circle] {$\sigma_{\bs{w}^\td}^2$};
  \node at (9.3,-1.9) [circle] {$\mu_{\bs{w}^{\tads,a}},\sigma_{\bs{w}^{\tads,a}}^2$};
  \node at (6.3,-1.9) [circle] {$\bs{\mu}_{\bs{V}^{\tads,a}},\sigma_{\bs{V}^{\tads,a}}^2\bs{I}$};
  \node at (4,-2) [circle] (uvr) {};
  \node at (2,-2) [circle] (uwr) {};
  \node at (8.8,-2) [circle] (uwa) {};
  \node at (6,-2) [circle] (uVa) {};
  \node at (4,-4) [circle,draw] (Vr) {$\bs{v}^{\tads}_i$};
  \node at (2,-4) [circle,draw] (wr) {$w^{\tads}_i$};
  \node at (8.8,-6) [circle,draw] (wa) {$w^{\tads,a}_l$};
  \node at (6,-4) [circle,draw] (Va) {$\bs{v}^{\tads,a}_l$};
  \node at (4,-6) [circle,draw] (xr) {$\bs{x}^{\tads}$};
  \node at (6,-6) [circle,draw] (xa) {$\bs{x}^{\tads,a}$};
  \node at (2,-8) [circle,draw] (w0r) {$w_0^{\tads}$};
  \node at (4,-8) [circle,draw] (yr) {$y^{\tads}$};
  \path[every node/.style={font=\sffamily\small}]
    (uvl) edge (Vl)
    (uwl) edge (wl)
    (uvr) edge (Vr)
    (uwr) edge (wr)
    (uwa) edge (wa)
    (uVa) edge (Va)
    (Vl) edge [bend left=45] (Vr)
    		 edge (yl)
	(wl) edge [bend left=45] (wr)
    		 edge (yl)
    	(xl) edge (yl)
    	(w0l)edge (yl)	
	(Vr) edge [bend right=25] (yr)
	(wr) edge (yr)
	(wa) edge (yr)
	(Va) edge (yr)
    	(xr) edge (yr)
    	(xa) edge (yr)
    	(w0r)edge (yr)
;
\draw (-0.5,-5.1) rectangle (-3,-9);
\draw (-0.3,-3) rectangle (-5,-7);
\draw (3,-5.1) rectangle (7,-9);
\draw (0.3,-3) rectangle (4.9,-7);
\draw (10,-3) rectangle (5.1,-7);
\node at(-4,-6.7) [circle] {$I^{\tweb} + J^{\tweb}$};
\node at(-0.8,-8.7) [circle] {$|D^{\tweb}|~~$};
\node at(1.1,-6.7) [circle] {$I^{\tads} +J^{\tads}$};
\node at(9.6,-6.7) [circle] {$L^{\tads}$};
\node at(6.7,-8.7) [circle] {$|D^{\tads}|~~$};
\node at(5.7,-9.4) [circle] {CTR Task};
\node at(-1.8,-9.4) [circle] {CF Task};
\end{tikzpicture}
\end{center}
\vspace{-35pt}
\caption{Graphic model of transferred factorisation machines.}\vspace{-15pt}
\label{graphmodel}
\end{figure}

The graphic representation for the proposed \emph{transferred factorisation machines} is depicted in Figure~\ref{graphmodel}. It illustrates the relationship among model parameters and observed data. The left part is for the CF task: $\bs{x}^{\tweb}$, $w_0^{\tweb}$, $\bs{w}^{\tweb}$ and $\bs{V}^{\tweb}$ work together to infer our CF task target $y^{\tweb}$, i.e., whether the user would visit a specific publisher or not. The right part illustrates the CTR task. Corresponding to CF task, $\bs{w}^{\tads}$ and $\bs{V}^{\tads}$ here represent user and publisher features' weights and latent vectors, while $\bs{w}^{\tads,a}$ and $\bs{V}^{\tads,a}$ are separately depicted to represent ad features' weights and latent vectors. All these factors work together to predict CTR task target $y^{\tads}$, i.e., whether the user would click the ad or not. On top of that, for each (user or publisher) feature $i$ of the CF task, its weight $w^{\tweb}_i$ and latent vector $\bs{v}^{\tweb}_i$ act as a prior of the counterparts $w^{\tads}_i$ and $\bs{v}^{\tads}_i$ in CTR task while learning the model.


Considering the datasets of the two tasks might be seriously unbalanced, we choose to focus on the \emph{averaged} log-likelihood of generating each data instance from the two tasks. In addition, we add a hyperparameter $\alpha$ for balancing the task relative importance.
As such, the joint conditional likelihood in Eq.~(\ref{eq:map}) is written as
\begin{align}
\Big[ \prod_{(\bs{x}^{\tweb},y^{\tweb}) \in D^{\tweb}} P(y^{\tweb}|\bs{x}^{\tweb}; \Theta) \Big]^{\frac{\alpha}{|D^{\tweb}|}} \cdot \Big[ \prod_{(\bs{x}^{\tads},y^{\tads}) \in D^{\tads}} P(y^{\tads}|\bs{x}^{\tads}; \Theta) \Big]^{\frac{1 - \alpha}{|D^{\tads}|}} \label{eq:joint-likelihood}
\end{align}
and its log form is
\begin{align}
&\frac{\alpha}{|D^{\tweb}|} \sum_{(\bs{x}^{\tweb},y^{\tweb}) \in D^{\tweb}} \Big[ y^{\tweb} \log \hat{y}_{u,p}^{\tweb} + (1 - y^{\tweb}) \log (1 - \hat{y}_{u,p}^{\tweb}) \Big] \nonumber \\
&~~~~~~~~~ + \frac{1 - \alpha}{|D^{\tads}|} \sum_{(\bs{x}^{\tads},y^{\tads}) \in D^{\tads}} \Big[ y^{\tads} \log \hat{y}_{u,p,a}^{\tads} + (1 - y^{\tads}) \log (1 - \hat{y}_{u,p,a}^{\tads})\Big].\label{eq:log-data}
\end{align}

Moreover, from the graphic model, the prior of model parameters can be specified as
\begin{align}
P(\Theta) =& P(\bs{w}^{\tweb})P(\bs{V}^{\tweb})P(\bs{w}^{\tads}|\bs{w}^{\tweb})P(\bs{V}^{\tads}|\bs{V}^{\tweb})P(\bs{w}^{\tads,a})P(\bs{V}^{\tads,a}) \\
\log P(\Theta) =& \sum_i \log\mathcal{N}(w^{\tweb}_i;\mu_{\bs{w}^{\tweb}},\sigma_{\bs{w}^{\tweb}}^2) + \sum _i \log\mathcal{N}(\bs{v}^{\tweb}_i;\bs{\mu}_{\bs{V}^{\tweb}},\sigma_{\bs{V}^{\tweb}}^2\bs{I}) \nonumber\\
& + \sum_i \log\mathcal{N}(w^{\tads}_i;w^{\tweb}_i,\sigma_{\bs{w}^\td}^2)
 + \sum_i \log \mathcal{N}(\bs{v}^{\tads}_i;\bs{v}^{\tweb}_i,\sigma_{\bs{V}^\td}^2\bs{I}) \label{eq:log-theta} \\
& + \sum_l \log \mathcal{N}(w^{\tads,a}_l;\mu_{\bs{w}^{\tads,a}},\sigma_{\bs{w}^{\tads,a}}^2) + \sum_l \log \mathcal{N}(\bs{v}^{\tads,a}_l;\bs{\mu}_{\bs{V}^{\tads,a}},\sigma_{\bs{V}^{\tads,a}}^2\bs{I}).\nonumber
\end{align}

\subsection{Learning the Model}
Given the detailed implementations of the MAP solution (Eq.~(\ref{eq:map})) components in Eqs.~(\ref{eq:log-data}) and (\ref{eq:log-theta}), for each data instance $(\bs{x}, y)$, the gradient update of $\Theta$ is
\begin{align}
\Theta \leftarrow \Theta + \eta \Big( \beta \frac{\partial}{\partial\Theta}\log P(y|\bs{x}; \Theta) + \frac{\partial}{\partial\Theta}\log P(\Theta) \Big),
\end{align}
where $P(y|\bs{x}; \Theta)$ is as Eqs.~(\ref{eq:web-pred}) and (\ref{eq:ads-pred}) for $(\bs{x}^{\tweb}, y^{\tweb}) \in D^{\tweb}$ and $(\bs{x}^{\tads}, y^{\tads}) \in D^{\tads}$, respectively; $\eta$ is the learning rate; $\beta$ is the instance weight parameter depending on which task the instance belongs to, as given in Eq.~(\ref{eq:log-data}). The detailed gradient for each specific parameter can be calculated routinely and thus are omitted here due to the page limit.

\section{Experiments} \label{sec:exp}

\subsection{Dataset}
Our experiments are conducted based on a real-world dataset provided by Adform, a global digital media advertising technology company based in Copenhagen, Denmark. It consists of two weeks of online display ad logs across different campaigns during March 2015. Specifically, there are 42.1M user domain browsing events and 154.0K ad display/click events. To fit the data into the joint model, we group useful data features into three categories: user features $\bs{x}^u$ (\ft{user\_cookie}, \ft{hour}, \ft{browser}, \ft{os}, \ft{user\_agent} and \ft{screen\_size}), publisher features $\bs{x}^p$ (\ft{domain}, \ft{url}, \ft{exchange}, \ft{ad\_slot} and \ft{slot\_size}), ad features $\bs{x}^a$ (\ft{advertiser} and \ft{campaign}). Detailed unique value numbers for each attribute are given as below.

{
\centering
\vspace{3pt}
\resizebox{\columnwidth}{!}{
\begin{tabular}{r|ccccccc}
\textsc{Attribute} & \ft{user\_cookie~} & \ft{hour~} & \ft{browser~} & \ft{os~} & \ft{user\_agent~} & \ft{screen\_size~} & \ft{domain}\\
\textsc{Unique number} & 4,180,170 & 24 & 71 & 37 & 29,488 & 118 & 38,495\\ \hline
\textsc{Attribute} & \ft{url} & \ft{exchange} & \ft{position} & \ft{size} & \ft{advertiser} & \ft{campaign}\\
\textsc{Unique number} & 1,100,523 & 140 & 3 & 55 & 486 & 2,665\\
\end{tabular}
}
~\\
}

In order to perform stable knowledge transfer, we have down-sampled the negative instances to make the ratio of positive over negative instances as 1:5.\footnote{It is common to perform negative down sampling to balance the labels in ad CTR estimation \cite{he2014practical}. Calibration methods \cite{caruana2006empirical} are then leveraged to eliminate the model bias.}


\subsection{Experiment Protocol}
We conduct a two-stage experiment to verify the effectiveness of our proposed models. First, in a very clean setting, we only focus on \ft{user\_cookie} and \ft{domain} to check whether the knowledge of users' behaviour on webpage browsing can be transferred to model their behaviour on clicking the ads in these webpages. Second, we start to append various features in the first setting to observe the performance change and check which features lead to better transfer learning. Specifically, we try appending a single side feature into the baseline setting: 1. appending user feature $\bs{x}^u$, 2. appending publisher feature $\bs{x}^p$, 3. appending ad feature $\bs{x}^a$. Finally, all features are added into the model to perform the transfer learning.

For each experiment stage, there are three datasets: \textsc{CF dataset} ($D^{\tweb}$), \textsc{CTR dataset} ($D^{\tads}$) and \textsc{Joint dataset}  ($D^{\tweb},D^{\tads}$). Each dataset is split into two parts: the first week data as training data and the second one as test data.



\subsection{Evaluation Metrics}

To evaluate the performance of proposed model, area under the ROC curve (AUC) \cite{graepel2010web} and root mean square error (RMSE) \cite{koren2009matrix} are adopted as performance metrics. As we focus on ad click prediction performance improvement, we only report the performance of the CTR estimation task.


\subsection{Compared Models}
We implement the following models for experimental comparison. 
\begin{itemize}
\item \tsf{Base}: This baseline model only considers the ad CTR task, without any transfer learning. The parameters are learned by $\max_{\Theta} \prod_{(\bs{x}^{\tads},y^{\tads})\in D^{\tads}} P(y^{\tads}|\bs{x}^{\tads};\Theta) P(\Theta)$.
\item \tsf{Disjoint}: This method performs a knowledge transfer in a disjoint two-stage fashion. First, we train the CF task model to get the parameters $\bs{w}^{\tweb}$ and $\bs{V}^{\tweb}$ by $\max_{\Theta} \prod_{(\bs{x}^{\tweb},y^{\tweb})\in D^{\tweb}} P(y^{\tweb}|\bs{x}^{\tweb};\Theta) P(\Theta)$. Second, with the CF task parameters fixed, we train the CTR task using Eqs.~(\ref{eq:log-data}) and (\ref{eq:log-theta}). Note that $\alpha$ in Eq.~(\ref{eq:log-data}) is still a hyperparameter for this method.
\item \tsf{DisjointLR}: The transfer learning model proposed in \cite{dalessandro2014scalable} is considered as state-of-the-art transfer learning methods in display advertising. In this work, both source and target tasks adopt logistic regression as a behaviour prediction model, which uses the linear model to minimise the logistic loss from each observation sample:
    \begin{align}
    \mathcal{L}_{\bs{w}}(\bs{x}, y) = -y \log \sigma (\langle\bs{w},\bs{x}\rangle) - (1-y)\log (1-\sigma (\langle \bs{w},\bs{x} \rangle)).
    \end{align}
    In our context of regarding the CF task as source task and CTR task as target task, the learning objectives are listed below:
    \begin{align}
    \textsc{CF task}: \overset{*}{\bs{w}^{\tweb}} &= \arg \min_ {\bs{w}^{\tweb}}\sum_{(\bs{x}^{\tweb},y^{\tweb})\in D^{\tweb}} \mathcal{L}_{\bs{w}^{\tweb}}(\bs{x}^{\tweb}, y^{\tweb}) + \lambda ||\bs{w}^{\tweb}||^2_2 \label{eq:lr-1}\\
    \textsc{CTR task}: \overset{*}{\bs{w}^{\tads}} &= \arg \min_ {\bs{w}^{\tads}}\sum_{(\bs{x}^{\tads},y^{\tads}) \in D^{\tads}} \mathcal{L}_{\bs{w}^{\tads}}(\bs{x}^{\tads}, y^{\tads}) + \lambda || \bs{w}^{\tads} - \overset{*}{ \bs{w}^{\tweb}} ||^2_2. \label{eq:lr-2}
    \end{align}
    Besides the difference between the linear LR and non-linear FM, this method is a two-stage learning scheme, where the first stage Eq.~(\ref{eq:lr-1}) is disjoint with the second stage Eq.~(\ref{eq:lr-2}). Thus we denoted it as \tsf{DisjointLR}.
\item \tsf{Joint}: Our proposed model, as summarised in Eq.~(\ref{eq:map}), which performs the transfer learning when jointly learning the parameters on the two tasks.
\end{itemize}

\subsection{Result}

\begin{figure}[t]
\centering
\vspace{-30pt}
\includegraphics[width=1.05\columnwidth]{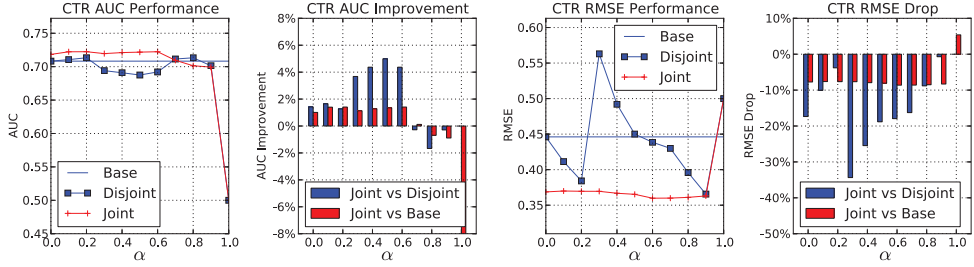}\vspace{-8pt}
\caption{Performance improvement with basic setting.}\label{fig:basic-perf}
\vspace{-7pt}
\end{figure}

\textbf{Basic Setting Performance.}
Figure~\ref{fig:basic-perf} presents the AUC and RMSE performance of \tsf{Base}, \tsf{Disjoint} and \tsf{Joint} and the improvement of \tsf{Joint} against the hyperparameter $\alpha$ in Eq.~(\ref{eq:log-data}) based on the basic experiment setting. As can be observed clearly, for a large region of $\alpha$, i.e., $[0.1, 0.7]$, \tsf{Joint} consistently outperforms the baselines \tsf{Base} and \tsf{Disjoint} on both AUC and RMSE, which demonstrates the effectiveness of our model to transfer knowledge from webpage browsing data to ad click data. Note that when $\alpha=0$, the CF side model $\bs{w}^{\tweb}$ does not learn but \tsf{Joint} still outperforms \tsf{Disjoint} and \tsf{Base}. This is due to the different prior of $\bs{w}^{\tads}$ and $\bs{V}^\tads$ in \tsf{Joint} compared with those of \tsf{Disjoint} and \tsf{Base}. In addition, when $\alpha=1$, i.e., no learning on CTR task, the performance of \tsf{Joint} reasonably gets back to initial guess, i.e., both AUC and RMSE are 0.5.

\begin{table}[t]\vspace{-10pt}
\centering
\caption{Overall AUC performance: \tsf{DisjointLR} vs \tsf{Joint}.} \label{t:lr-impr}
\vspace{-7pt}
\begin{tabular}{ccc}
\tsf{DisjointLR} & \tsf{Joint} & \ft{Improvement}\\\hline
68.44\% & 72.18\% & 5.46\% \\
\end{tabular}
\vspace{-0pt}
\end{table}

Table \ref{t:lr-impr} shows the transfer learning performance comparison between \tsf{Joint} and the state-of-the-art \tsf{DisjointLR} with both models setting optimal hyperparameters. The improvement of \tsf{Joint} over \tsf{DisjointLR} indicates the success of 1) the joint optimisation on the two tasks to perform knowledge transfer and 2) the non-linear factorisation machine relevance model on catching feature interactions.


\vspace{5pt}
\noindent \textbf{Appending Side Information Performance.}
From the \tsf{Joint} model as in Eq.~(\ref{eq:log-data}) we see when $\alpha$ is large, e.g., 0.8, the larger weight is allocated on the CF task to optimise the joint likelihood. As such, if a large-value $\alpha$ leads to the optimal CTR estimation performance, it means the transfer learning takes effect. With such method, we try adding different features into the \tsf{Joint} model to obtain the optimal hyperparameter $\alpha$ leading to the highest AUC to check whether a certain feature helps transfer learning. On the contrary, if a low-value or 0 $\alpha$ leads to the optimal performance of \tsf{Joint} model when adding a certain feature, it means such feature has no effect of performing transfer learning.

\begin{figure}[t]
\vspace{-30pt}
\centering
\includegraphics[width=1.05\columnwidth]{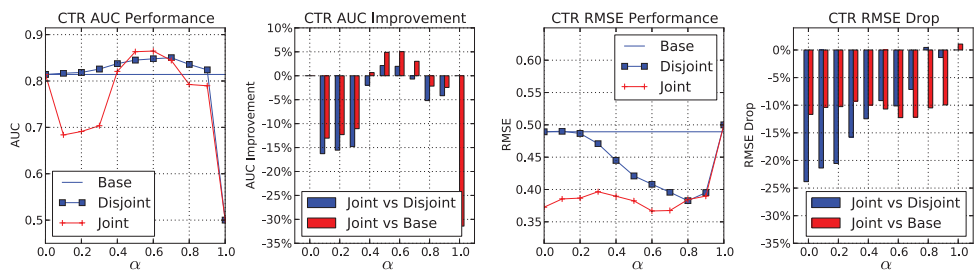}\vspace{-8pt}
\caption{Performance improvement with different side information.}\label{fig:all-feat-perf}
\vspace{-7pt}
\end{figure}

Table~\ref{tab:side-info-perf} collects the AUC improvement of the \tsf{Joint} model for the conducted experiments. We observe that user browsing \ft{hour}, ad slot \ft{position} in the webpage are the most valuable features that help transfer learning, while the user \ft{screen size} does not bring any transfer value. When adding all these features into \tsf{Joint} model, the optimal $\alpha$ is around 0.5 for AUC improvement and 0.6 for RMSE drop (see Figure~\ref{fig:all-feat-perf}), which means these features along with the basic user, webpage IDs provide an overall positive value of knowledge transfer from webpage browsing behaviour to ad click behaviour.

\begin{table*}[t]
\centering
\scriptsize
\caption{CTR task performance}\label{tab:side-info-perf}
\vspace{-5pt}
\begin{tabular}{|l||p{0.5cm}|p{1cm}|p{1cm}|p{1cm}||p{0.5cm}|p{1cm}|p{1cm}|p{1cm}|}\hline
&\multicolumn{4}{c||}{\tsf{Joint} vs \tsf{Disjoint}}& \multicolumn{4}{c|}{\tsf{Joint} vs \tsf{Base}}\\\cline{2-9}
&$\overset{*}{\alpha}$ & AUC Lift & \tsf{Joint} AUC & \tsf{Disjoint} AUC & $\overset{*}{\alpha}$ & AUC Lift & \tsf{Joint} AUC & \tsf{Base} AUC(\%)\\ \hline
\textsc{Basic Setting} & 0.5 & 3.43\% & 72.18\% & 68.75\% & 0.2 & 1.41\% & 72.24\% & 70.83\% \\\hline
+ $\bs{x}^u$: \ft{hour} & 0.8 & 2.44\% & 89.35\% & 86.91\% & 0.6 & 1.99\% & 89.35\% & 87.36\% \\
+ $\bs{x}^u$: \ft{browser} & 0.0 & 7.92\% & 76.36\% & 68.44\% & 0.2 & 8.08\% & 76.52\% & 68.44\%\\
+ $\bs{x}^u$: \ft{os} & 0.1 & 6.66\% & 76.86\% & 70.2\% & 0.1 & 6.71\% & 76.86\% & 70.15\%\\
+ $\bs{x}^u$: \ft{user\_agent} & 0.0 & 2.57\% & 67.12\% & 64.55\% & 0.8 & 4.31\% & 68.86\% & 64.55\%\\
+ $\bs{x}^u$: \ft{screen\_size} & 0.0 & 9.39\% & 76.43\%  & 67.04\% & 0.0 & 9.39\% & 76.43\% & 67.04\%
\\\hline
+ $\bs{x}^p$: \ft{exchange} & 0.6 & 1.56\% & 66.80\% & 65.24\% & 0.0 & 0.64\% & 68.49\% & 67.85\%\\
+ $\bs{x}^p$: \ft{url} & 0.3 & 11.9\% & 66.56\% & 54.66\% & 0.0 & 11.55\% & 69.36\% & 57.81\%\\
+ $\bs{x}^p$: \ft{position} & 0.6 & 2.63\% & 66.89\% & 64.26\% & 0.4 & 0.69\% & 67.14\% & 66.45\%
\\\hline
+ $\bs{x}^a$: \ft{advertiser} & 0.4 & 2.39\% & 84.98\% & 82.59\% & 0.5 & 0.87\% & 85.07\% & 84.20\%\\
+ $\bs{x}^a$: \ft{campaign} & 0.2 & 1.29\% & 85.81\% & 84.52\% & 0.1 & 0.48\% & 85.91\% & 85.43\% \\
+ $\bs{x}^a$: \ft{size} & 0.0 & 0.59\% & 69.16\% & 68.57\% & 0.0 & 0.59\% & 69.16\% & 68.57\% \\ \hline
+ \textsc{all features} & 0.5 & 6.91\% & 88.32\% & 81.41\% & 0.6 & 6.91\% & 88.32\% & 81.41\% \\\hline
\end{tabular}
\vspace{-10pt}
\end{table*}

\section{Conclusion}
In this paper, we proposed a transfer learning framework with factorisation machines to build implicit look-alike models on user ad click behaviour prediction task with the knowledge successfully transferred from the rich data of user webpage browsing behaviour. The major novelty of this work lies in the joint training on the two tasks and making knowledge transfer based on the non-linear factorisation machine model to build the user and other feature profiles. Comprehensive experiments on a large-scale real-world dataset demonstrated the effectiveness of our model as well as some insights of detecting which specific features help transfer learning. In the future work, we plan to explore on the user profiling utilisation based on the learned latent vector for each user. We also plan to extend our model to cross-domain recommendation problems.

\section*{Acknowledgement}
We would like to thank Adform for allowing us to use their data in experiments. We would also like to thank Thomas Furmston for his feedback on the paper. Weinan thanks Chinese Scholarship Council for the research support.

\bibliographystyle{splncs03}
\bibliography{ecir18-zhang}

\end{document}